\documentclass[11pt]{article} 

\usepackage{styles/rldmsubmit}
\usepackage{palatino}
\usepackage[authoryear,sort&compress,round]{natbib}
\usepackage{microtype}
\usepackage{graphicx}
\usepackage{subfigure}
\usepackage{centernot}
\usepackage{cancel}
\usepackage{booktabs}
\usepackage{hyperref}
\usepackage{dashrule}

\usepackage{cancel}
\usepackage{amsmath}
\usepackage{amssymb}
\usepackage{mathtools}
\usepackage{amsthm}
\usepackage{scalerel}
\usepackage{enumitem}

\usepackage[capitalize,noabbrev,nameinlink]{cleveref}
\creflabelformat{equation}{#2#1#3}
\usepackage{csquotes}
\usepackage{multirow}
\usepackage{mathrsfs}
\usepackage{mdframed}
\usepackage{bm}
\usepackage{bbm,tipa}
\usepackage{mathbbol}
\DeclareSymbolFontAlphabet{\mathbb}{AMSb}
\DeclareSymbolFontAlphabet{\mathbbl}{bbold}

\usepackage{xcolor}
\definecolor{dred}{RGB}{153,80,43}
\definecolor{dblue}{RGB}{0,114,178}
\hypersetup{
    colorlinks=true,
    breaklinks=true,
    urlcolor=dblue, 
    linkcolor=dred, 
    citecolor=dblue
}
\usepackage{caption}

\def\environment{e}
\def\env{\environment}

\newcommand\reaches{%
  \mathrel{\ooalign{\hss$\rightsquigarrow$\hss\kern-1.45ex\raise1.0ex\hbox{{\scaleobj{0.55}{\env}}}\hspace{4pt}}}}
\newcommand\cnot[1]{%
  \mathrel{\ooalign{\hfil$#1$\hfil\cr\hfil$/$\hfil\cr}}}
\newcommand\notreaches{%
  \mathrel{\ooalign{\hss$\cnot\rightsquigarrow$\hss\kern-1.45ex\raise1.0ex\hbox{{\scaleobj{0.55}{\env}}}\hspace{4pt}}}}

\newcommand\alwaysreach{%
  \mathrel{\ooalign{\hss$\square$\hss\kern-0.22ex\hbox{{$\reaches$}}}}}

\newcommand\generates{%
 \mathrel{\ooalign{\hss$\vdash$\hss\kern-0.65ex\raise0.9ex\hbox{{\scaleobj{0.55}{\env}}}}}}

\usepackage{calligra}
\DeclareMathAlphabet{\mathcalligra}{T1}{calligra}{m}{n}

\newtheorem{claim}{Claim}
\newtheorem*{theorem*}{Theorem}
\newtheorem*{axiom*}{Axiom}

\newtheorem*{definition*}{Definition}

\numberwithin{equation}{section}
\numberwithin{theorem}{section}
\numberwithin{definition}{section}
\numberwithin{conjecture}{section}

\newmdenv[
  topline=false,
  bottomline=false,
  rightline = false,
  leftmargin=8pt,
  rightmargin=0pt,
  skipabove=16pt, 
  innertopmargin=0pt,
  innerbottommargin=0pt
]{innerproof}

\usepackage{etoolbox}
\makeatletter
\patchcmd{\NAT@test}{\else \NAT@nm}{\else \NAT@hyper@{\NAT@nm}}{}{}
\makeatother

\usepackage{tikz}

\usepackage[textsize=tiny]{todonotes}

\newcommand{\creflink}[1]{\hyperref[#1]{\textcolor{blue}{\cref{#1}}}}

\newcommand{\ci}{\perp\kern-5pt\perp}

\title{Agency Is Frame-Dependent}

\author{
David Abel \\
Google DeepMind \\
\And
Andr{\' e} Barreto \\
Google DeepMind \\
\And
Michael Bowling \\
Amii, University of Alberta \\
\And
Will Dabney \\
Google DeepMind \\
\AND
Shi Dong \\
Google DeepMind \\
\And
Steven Hansen \\
Google DeepMind \\
\And
Anna Harutyunyan \\
Google DeepMind \\
\And
Khimya Khetarpal \\
Google DeepMind \\
\AND
Clare Lyle \\
Google DeepMind \\
\And
Razvan Pascanu \\
Google DeepMind \\
\And
Georgios Piliouras \\
Google DeepMind \\
\And
Doina Precup \\
Google DeepMind \\
\AND
Jonathan Richens \\
Google DeepMind \\
\And
Mark Rowland \\
Google DeepMind \\
\And
Tom Schaul \\
Google DeepMind \\
\And
Satinder Singh \\
Google DeepMind \\
}

\begin{document}
\maketitle

\begin{abstract}
%
Agency is a system's capacity to steer outcomes toward a goal, and is a central topic of study across biology, philosophy, cognitive science, and artificial intelligence.
%
Determining if a system exhibits agency is a notoriously difficult question: \cite{dennett1989intentional}, for instance, highlights the puzzle of determining which principles can decide whether a rock, a thermostat, or a robot each possess agency. 
%
We here address this puzzle from the viewpoint of reinforcement learning by arguing that agency is fundamentally frame-dependent: \textit{Any measurement of a system's agency must be made relative to a reference frame}. 
%
We support this claim by presenting a philosophical argument that each of the essential properties of agency proposed by \cite{barandiaran2009defining} and \cite{moreno2018minimal} are themselves frame-dependent. 
%
We conclude that any basic science of agency requires frame-dependence, and discuss the implications of this claim for reinforcement learning.
\end{abstract}

\keywords{
Agency, Philosophy of Reinforcement Learning
}

\acknowledgements{The authors would like to thank Kim Stachenfeld and Vlad Mnih for their thoughtful comments on a draft of the paper.}

\startmain
\section{Introduction}

%
Reinforcement learning (RL) involves learning or decision making over time to achieve a goal. Agents are often taken as the primary vehicles that carry out this learning and decision making, and as such have long been an essential element of RL. Moreover, \textit{agency} is the lifeblood of an agent---agency is the capacity that endows a given system with the status of agent-hood. Thus, agency also stands as one of the elemental concepts of RL. 
%
In \textit{The Evolution of Agency}, \citet{tomasello2022evolution} makes an even stronger case for the role of agency in psychology:
\begin{quote}
    Every scientific discipline begins with a proper domain, a first principle ... In psychology, depending on one’s theoretical predilections, that proper domain or first principle might be either behavior or mentality. But my preferred candidate would be agency, precisely because agency is the organizational framework within which both behavioral and mental processes operate.  (p. 134, \citeauthor{tomasello2022evolution}, \citeyear{tomasello2022evolution}).
\end{quote}
%
Following similar reasoning to Tomasello, we take it as essential that the science of RL is borne from an understanding not just of intelligence, learning, and decision making, but also of agency. To this end, we here investigate a fundamental question about agency through the lens of RL: is agency an invariant, measurable property of an input-output system, or does it vary depending on other independent commitments? We draw from several distinct results within the RL literature to arrive at the conclusion that agency is fundamentally frame-dependent.

%
\begin{figure}[b!]
    \centering
    
    %
    \subfigure[\label{subfig:agency_def}A Four-Part Account of Agency]{\includegraphics[width=0.3\textwidth]{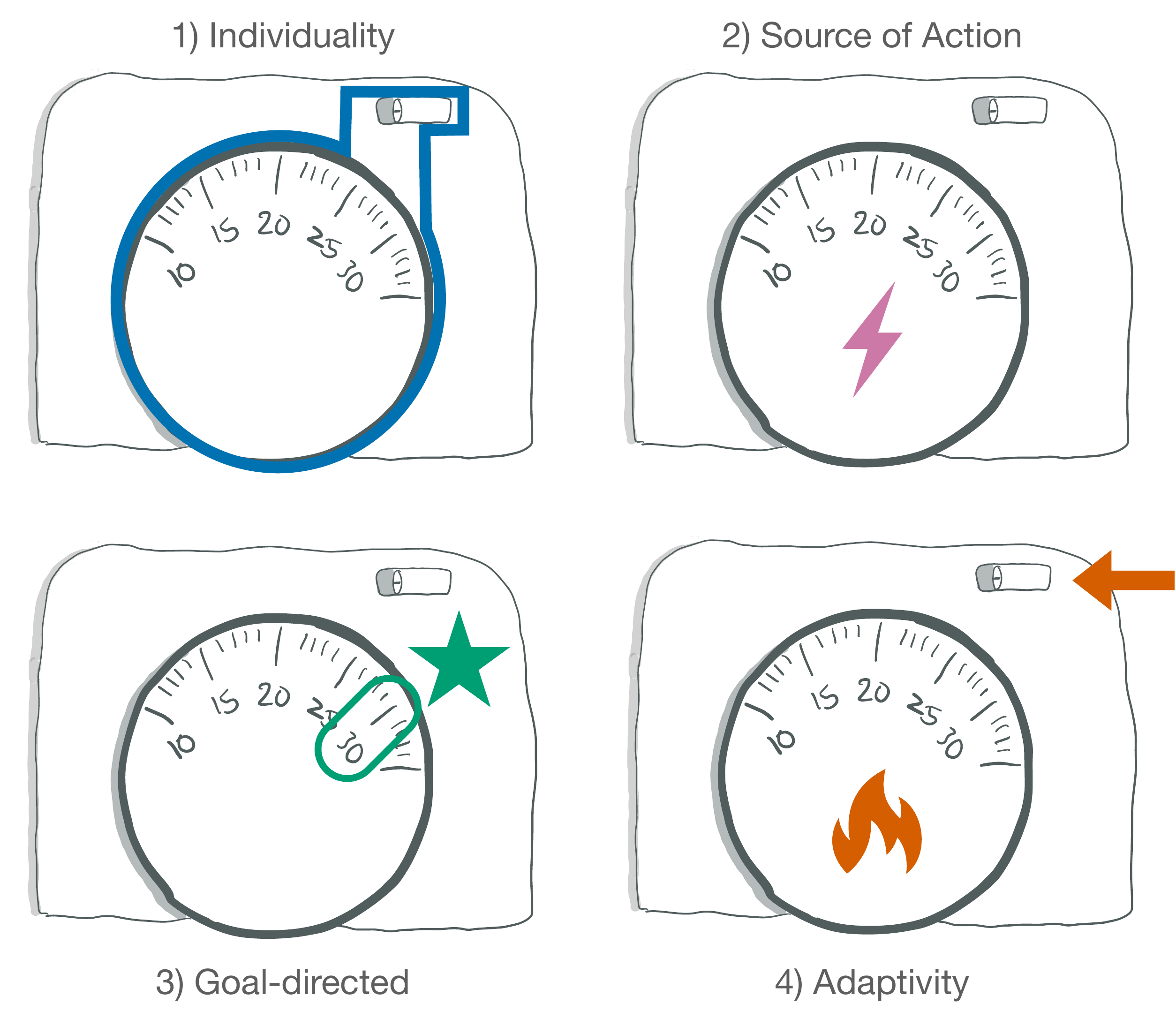}} \hspace{8mm}
    %
    %
    %
    \subfigure[\label{subfig:frames_example}Frame-Dependence]{\includegraphics[width=0.5\textwidth]{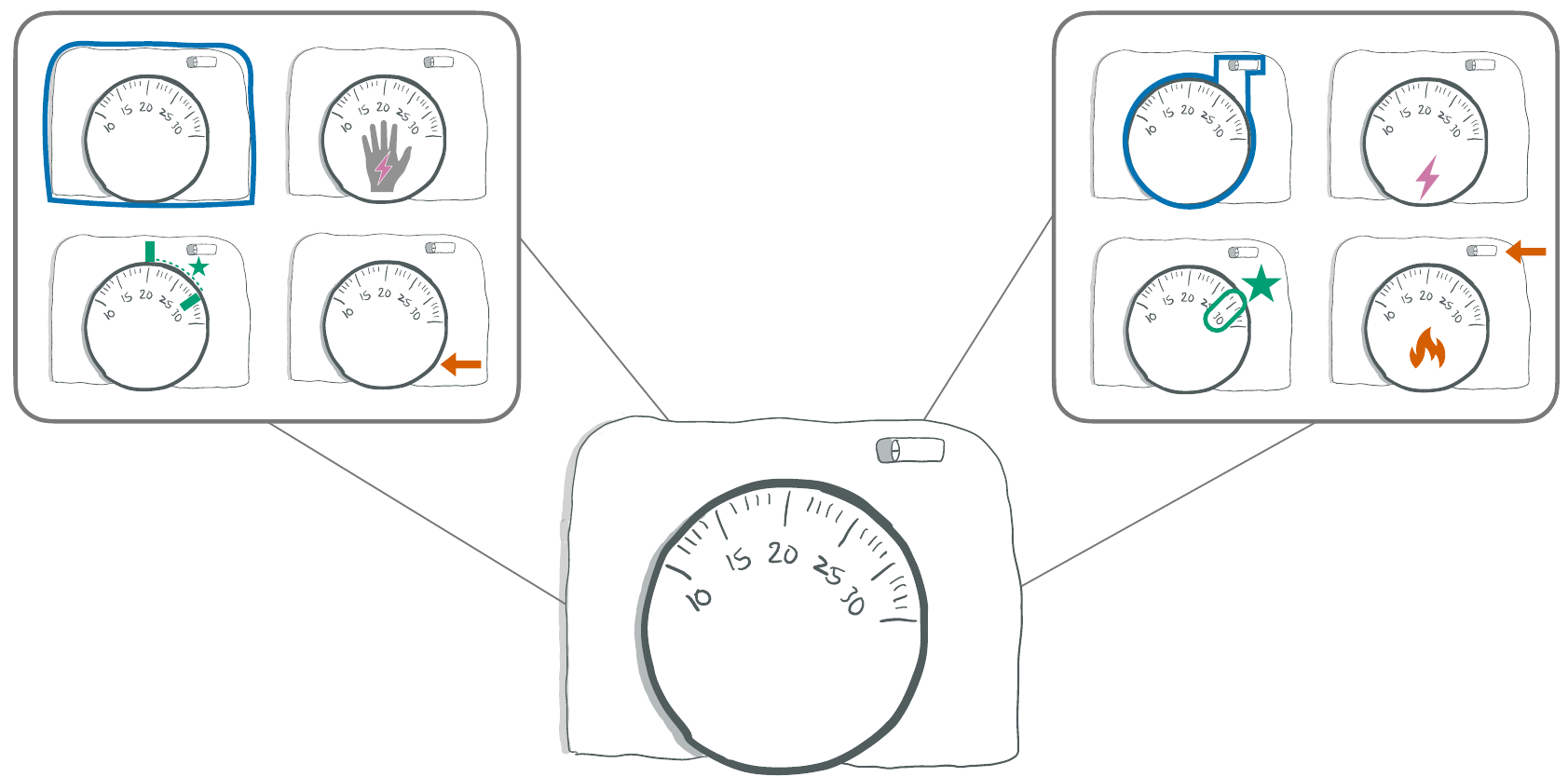}}
    
    %
    \caption{(Left) A four-part account of agency due largely to \cite{barandiaran2009defining}: a system such as a thermostat has agency if it has (1) a boundary, (2) is the source of its own actions, (3) has a goal, and (4) adaptively selects outputs based on inputs to pursue this goal. (Right) Our main claim: A determination of the agency of a system, such as a thermostat, is dependent on a choice of reference frame. The two reference frames depicted make different commitments about how we measure the four essential conditions of agency. For example, we could draw the boundary around our thermostat in several different ways, or understand the goal of the thermostat in different ways.} 
    
    \label{fig:frame_dependence}
\end{figure}

%
\paragraph{Agency.} Typical views of agency across biology \citep{ball2023organisms}, complex systems \citep{moreno2005agency,moreno2018minimal}, and philosophy \citep{dretske1999machines,barandiaran2009defining} roughly define the concept as an input-output system's capacity to steer outcomes toward a goal. 
%
We build around one canonical definition of agency developed by \cite{barandiaran2009defining} that we present in four parts. 
%
First, to have agency, the system must be \textit{individual}; it has a boundary that separates it as an independent entity from its surroundings. 
%
Second, once we have chosen a boundary, the system must be the source of its own action. 
%
Third, the system has some goals or norms that regulate its interactions with the environment. 
%
Fourth, the system steers its experience in light of these goals. 
%
%
We take these four conditions to be a reasonable starting point for any account of agency, summarized in \cref{subfig:agency_def}. %
For the purposes of our main argument, we do not take debates about the precise definition of agency to have a significant impact on our conclusion---we anticipate that regardless of how the semantics of "agent" or "agency" are worked out, the need for frame dependence will remain. For instance, Barandiaran et al. combine normativity and adaptivity into one property: this definition is perfectly valid, but will still admit frame-dependence. For further debates on this topic, see work by \cite{moreno2018minimal}, \cite{ball2023organisms}, or \cite{dretske1999machines}.

%
\paragraph{The Puzzle of Agency.} A fundamental puzzle then arises: which systems can be said to have agency? \cite{dennett1989intentional} considers the cases of a rock rolling down a hill and a thermostat modulating the temperature of a room---in what sense do these two systems possess agency, and to what extent? And, the more critical scientific question: what principles can we turn to in order to determine whether each of these systems possess agency? Barandiaran et al. stipulate that a system possesses agency if the four conditions are present within the system. If any one of them is missing, the system lacks agency. In this way, agency is taken to be binary, though naturally there is room for developing a non-binary account.

\section{Agency is Frame-Dependent}

%
\paragraph{Main Claim.} We here address this puzzle by arguing that the attribution of agency to a system is \textit{fundamentally dependent on a choice of reference frame}. That is, the agency of any system is \textit{relative} in the sense that it depends on arbitrary commitments that we collectively call a reference frame. For example, one such commitment is whether the system is meaningfully pursuing a goal; depending on how we codify what counts as meaningful goal-pursuit, the system will either be understood as having agency or not.  
%
We support this main claim by illustrating that each of the four properties of agency are themselves \textit{relative} to a choice of some reference object or commitment---that is, reaching a conclusion about whether a given system possesses each property requires an independent commitment whose choice is arbitrary.
%
At present, our definitions, claims, and arguments are purely philosophical, though we note that a rigorous presentation of this perspective is a natural and fruitful direction for future work.

%
\paragraph{What is a Reference Frame?} An agent reference frame is a collection of these four commitments that allow us to determine whether a system has each of the four properties. That is, a frame must include (1) a boundary that decides what is internal to the agent and what is external; (2) a reference object such as a set of causal variables that allow for determination of whether the system is the cause of its action; (3) a principle for isolating whether the system is meaningfully pursuing a goal; and (4) a choice of what changes in behavior count as meaningful adaptation. There are many valid ways to formalise these components; a boundary could be a cut in a graph \citep{jiang2019value} or a Markov blanket \citep{friston2009reinforcement}. A precise mathematical construction of reference frames is a natural next step for further research. 

%
\paragraph{(1) Individuality.} In order to attribute agency to an entity, an observer first must determine which entity they are referring to. Establishing a boundary that separates this entity from its surroundings is a critical step in determining whether a system has agency. However, clearly identifying such a boundary is non-trivial. 
%
Neils Bohr gives the example of a person wielding a stick---depending on the activity, the stick could be taken as a part of the person's propensity for both action and observation (p. 93, \citeauthor{klein1967glimpses}, \citeyear{klein1967glimpses}). As \cite{harutyunyan2020what} notes, we might plausibly draw the boundary around the person and exclude or include the stick. 
%
In fact, as argued by \cite{clark1998extended}, \cite{jiang2019value}, and \cite{harutyunyan2020what}, there are often many ways we can establish a boundary that separates an environment and an agent in a meaningful way. Jiang considers the example of a model-free learning algorithm that is implemented using a neural network to process its observations. As Jiang points out, the boundary we choose to draw could include the pseudo-random number generator and all layers of this network, or only include the last few layers of the network. Proposition 10 by \cite{jiang2019value} (further discussed in Section 6.1), illustrates that many important quantities of an RL agent, such as the optimal policy or Bellman error, are \textit{boundary-dependent}. We summarize these points in the following claim.

%
\begin{claim}[Adapted from \citeauthor{jiang2019value}, \citeyear{jiang2019value} and \citeauthor{harutyunyan2020what}, \citeyear{harutyunyan2020what}]
Individuality is frame-dependent: Nearly all agents admit many plausible boundaries one could draw that separates them from their environment. Moreover, key quantities of an agent can change depending on which boundary is chosen. 
\end{claim}

%
Individuality might be seen as qualitatively different from the other properties, since it involves selecting \textit{which} system we would like to attribute agency to and what its inputs and outputs are. As such, it is natural to conclude that we recover a different agent depending on how we draw the boundary. However, our argument does not stand only on the frame-dependency of individuality---since we take all four key properties of agency to be frame-dependent, our argument remains agnostic to any perceived qualitative difference between them. 
%
%

%
\paragraph{(2) Source of Action.} Second, a system must be the source of its own action. In the terms of \cite{ball2023organisms}, this property reflects whether the system is "pushed around by its environment", or does the pushing itself. 
%
For instance, a wall being knocked over by a wrecking ball could be understood as taking the action of being knocked over. However, the source of this action (and the corresponding potential energy) did not ultimately originate in the wall, but rather in the wrecking ball and its operator. It might therefore be better to view the wall as lacking in the source of action. In contrast, a bird flapping its wings intuitively satisfies the condition since this action is best thought of as originating from within the bird. 

%
\cite{kenton2023discovering} recently develop a causal account that determines which entities in a causal model might be said to satisfy roughly this property. The difficulty, as Kenton et al. note, is that reaching a conclusion about the source of action in a causal model rests \textit{entirely} on the choice of causal variables. In this way, identifying whether a given subsystem originates its own action depends on an independent, unrelated choice: the causal variables. Kenton et al. state directly: "Note [discovering an agent in a causal model] is relative to a frame -- a choice of variables that appear in our causal model" (p. 2, \citeauthor{kenton2023discovering}, \citeyear{kenton2023discovering}). Following this reasoning, we claim that the source of action is also frame-dependent in that it depends on an arbitrary upstream commitment. %

%
\begin{claim}[Adapted from \citeauthor{kenton2023discovering}, \citeyear{kenton2023discovering}]
Source of action is frame-dependent: There exist cases with at least two plausible choices of causal variables where the former choice identifies an agent in the causal model, and the latter choice refutes the presence of an agent in the causal model.
\end{claim}

%
\paragraph{(3) Normativity.} Third, and perhaps most crucially, agency is about goal-directedness. The trouble is that every system with outputs can be understood \textit{as if} \citep{friedman1953essays} it is goal-directed. 

%
More concretely, every input-output system can be well-explained in terms of goal-directedness. For example, in the case of our thermostat, even a broken thermostat whose output always sets the temperature of a room to $20^\circ$ can be understood as having the goal "set the temperature to $20^\circ$". More trivially, a rock can be viewed as having the goal of rolling down a hill, or as having the goal of convincing all observers that it is a rock. What makes these meaningless, as opposed to meaningful goals? 
This challenge is reflected in one of the classical results of inverse RL first discussed by \cite{russell1998learning} and \cite{ng2000algorithms}: the zero reward function is always consistent with every system that outputs signals that are understood as decisions. In other words, reward is under-determined by behavior \citep{cao2021identifiability} without intervention \citep{amin2017repeated}. To overcome this challenge, a variety of approaches have explored the use of biases or other principles that constrain the space of viable reward functions, such as Occam's razor \citep{armstrong2018occam}, or the now standard approach invoking \textit{maximum entropy} proposed by \cite{ziebart2008maximum}. These upstream principles can in some cases rule out certain kinds of goal-directedness as uninteresting, or elevate others as meaningful. Hence, we must again invoke an additional principle to determine whether a given system is meaningfully goal-directed. One common approach is to ask how \textit{useful} it is to explain or understand the system in terms of goal-directedness, as first argued by \cite{dennett1989intentional}. 
In other words, to determine if a system \textit{meaningfully} has a goal, we require some extraneous commitments that again amount to a reference frame.

\begin{claim}
Normativity is frame-dependent: For nearly all cases of input-output systems, whether that system is meaningfully goal-directed depends on a reference point whose choice is arbitrary.
\end{claim}

%
\paragraph{(4) Adaptivity.} Fourth, agency is about \textit{adaptivity}, which captures whether a system's outputs are influenced by its inputs, and to what extent. 
%
%
In "On the Definition of Adaptivity", \cite{zadeh1963definition} suggests a form of frame-dependence: "...every system is adaptive with respect to [something] ... what matters is not whether [the system] is adaptive or not, but what ... it is adaptive [to]" (p. 470). Following this reasoning, adaptivity can be understood as frame-dependent in the sense that it depends on what is chosen as the relevant reference class used to reach determinations about a system's adaptivity. 
%
For example, in RL, we might ask whether a policy that maps each input state to an action is adaptive. On one reference frame, we might treat \textit{any} change in the output as adaptivity, while on another reference frame, we could view this policy as a fixed and non-adaptive function since it always chooses the same action every time it receives the same input.  \cite{abel2023crl} make this argument in a more general case and show that all policies can either be understood as adaptive, or not, depending on a reference class of meaningful changes in behavior. 
In other words, if we want to determine whether a system is adaptive, we need to first agree on the class of experience-influenced changes of behavior that count as adaptivity. Depending on this choice, a system will either be adaptive, or not. This choice then acts as a reference frame. 
%

%
\begin{claim}[Adapted from \citeauthor{zadeh1963definition}, \citeyear{zadeh1963definition} and Theorem 3.1 of \citeauthor{abel2023crl}, \citeyear{abel2023crl}]
Adaptivity is frame-dependent: For many input-output systems, there will exist at least two reference frames (and in most cases, many more) such that according to the first reference frame the system is adaptive, while according to the second, the system is not.
\end{claim}

%
In summary, reaching a determination about each of the four properties of agency requires reference to other fixed commitments that collectively comprise a reference frame. Since agency is simply the logical conjunction of these latter three properties conditioned on the choice of a boundary, then reaching a conclusion about agency itself must be made in reference to these commitments. In other words: agency is frame-dependent.

%

%

\section{Discussion}

%
We have here argued that agency is frame-dependent by illustrating the sense in which each of the four essential conditions of agency are themselves frame-dependent. We take this to have far reaching implications for disciplines that study agents and agency.

%
\paragraph{Intelligence and Agency.} The relationship between intelligence and agency is not yet well understood. For instance, does intelligence require agency, and vice versa? Depending on how this question is addressed, frame-dependence may have significant implications for our understanding of intelligence, in addition to other emergent properties of information processing systems. We believe that exploring this relationship through the lens of frame-dependence offers a new frontier for understanding central concepts of RL.

%
\paragraph{Reference Frames.} We stop short of presenting a rigorous mathematical definition of reference frames, as well as a formal proof of the frame-dependence of agency. A natural next step to further this line of work is to develop a precise definition of agent reference frames along with a formal proof supporting our main claim. We speculate that the building blocks to do so are already in place in the field, but have not gone through the careful work of developing the definitions and arguments more formally.

%
\paragraph{Choosing a Reference Frame.} How do we choose an appropriate reference frame? It is unclear which frame-selection principles are defensible, and what implications these principles carry for our study of agents. The Intentional Stance \citep{dennett1989intentional} asserts roughly that it is most \textit{useful} to understand certain systems as agents; taken to its natural conclusion, this could be formalised as a principle for selecting frames in terms of predictive or explanatory power. An important direction for further research will investigate, formalise, and compare different principles for selecting reference frames.


\paragraph{Marr's Levels, Dennett's Stances.} The proposal to adopt reference frames coheres with the perspectives of \cite{dennett1989intentional} as well as \cite{marr2010vision}, who each argue for understanding certain phenomena at different levels of abstraction. 
%
Marr argues that our attempts to understand a cognitive process such as vision can be cast through three distinct levels: the hardware, the algorithmic, and the computational. 
%
Dennett argues that there are distinct levels of abstraction we should adopt depending on the content of our study. Like Marr, the lowest-level of abstraction is the \textit{physical}, according to which we examine physical properties of a system. The second level is the design stance, according to which we examine how the system is designed to make predictions about its operation or purpose. The third is the intentional stance, according to which we study the content of minds themselves: of beliefs and desires.  %
Both Dennett's intentional stance and Marr's levels suggest that our study of concepts like agency needs to carefully calibrate in order to examine the right kinds of content, and reach the right kinds of conclusions. 
We suggest that reference frames may offer a path to connect physical substrates with more abstract propositions such as those related to agency.


\bibliographystyle{abbrvnat}
\bibliography{main}

\begin{thebibliography}{23}
\providecommand{\natexlab}[1]{#1}
\providecommand{\url}[1]{\texttt{#1}}
\expandafter\ifx\csname urlstyle\endcsname\relax
  \providecommand{\doi}[1]{doi: #1}\else
  \providecommand{\doi}{doi: \begingroup \urlstyle{rm}\Url}\fi

\bibitem[Abel et~al.(2023)Abel, Barreto, Roy, Precup, van Hasselt, and
  Singh]{abel2023crl}
D.~Abel, A.~Barreto, B.~V. Roy, D.~Precup, H.~van Hasselt, and S.~Singh.
\newblock A definition of continual reinforcement learning.
\newblock In \emph{Advances in Neural Information Processing Systems}, 2023.

\bibitem[Amin et~al.(2017)Amin, Jiang, and Singh]{amin2017repeated}
K.~Amin, N.~Jiang, and S.~Singh.
\newblock Repeated inverse reinforcement learning.
\newblock In \emph{Advances in Neural Information Processing Systems}, 2017.

\bibitem[Armstrong and Mindermann(2018)]{armstrong2018occam}
S.~Armstrong and S.~Mindermann.
\newblock Occam's razor is insufficient to infer the preferences of irrational
  agents.
\newblock In \emph{Advances in Neural Information Processing Systems}, 2018.

\bibitem[Ball(2023)]{ball2023organisms}
P.~Ball.
\newblock Organisms as agents of evolution.
\newblock \emph{John Templeton Foundation: West Conshohocken, PA, USA}, 2023.

\bibitem[Barandiaran et~al.(2009)Barandiaran, Di~Paolo, and
  Rohde]{barandiaran2009defining}
X.~E. Barandiaran, E.~Di~Paolo, and M.~Rohde.
\newblock Defining agency: Individuality, normativity, asymmetry, and
  spatio-temporality in action.
\newblock \emph{Adaptive Behavior}, 17\penalty0 (5):\penalty0 367--386, 2009.

\bibitem[Cao et~al.(2021)Cao, Cohen, and Szpruch]{cao2021identifiability}
H.~Cao, S.~Cohen, and L.~Szpruch.
\newblock Identifiability in inverse reinforcement learning.
\newblock In \emph{Advances in Neural Information Processing Systems}, 2021.

\bibitem[Clark and Chalmers(1998)]{clark1998extended}
A.~Clark and D.~Chalmers.
\newblock The extended mind.
\newblock \emph{Analysis}, 58\penalty0 (1):\penalty0 7--19, 1998.

\bibitem[Dennett(1989)]{dennett1989intentional}
D.~C. Dennett.
\newblock \emph{The intentional stance}.
\newblock MIT press, 1989.

\bibitem[Dretske(1999)]{dretske1999machines}
F.~I. Dretske.
\newblock Machines, plants and animals: the origins of agency.
\newblock \emph{Erkenntnis (1975-)}, 51\penalty0 (1):\penalty0 19--31, 1999.

\bibitem[Friedman(1953)]{friedman1953essays}
M.~Friedman.
\newblock \emph{Essays in positive economics}.
\newblock University of Chicago press, 1953.

\bibitem[Friston et~al.(2009)Friston, Daunizeau, and
  Kiebel]{friston2009reinforcement}
K.~J. Friston, J.~Daunizeau, and S.~J. Kiebel.
\newblock Reinforcement learning or active inference?
\newblock \emph{PloS One}, 4\penalty0 (7):\penalty0 e6421, 2009.

\bibitem[Harutyunyan(2020)]{harutyunyan2020what}
A.~Harutyunyan.
\newblock What is an agent?
\newblock
  \url{http://anna.harutyunyan.net/wp-content/uploads/2020/09/What_is_an_agent.pdf},
  2020.

\bibitem[Jiang(2019)]{jiang2019value}
N.~Jiang.
\newblock On value functions and the agent-environment boundary.
\newblock \emph{arXiv preprint arXiv:1905.13341}, 2019.

\bibitem[Kenton et~al.(2023)Kenton, Kumar, Farquhar, Richens, MacDermott, and
  Everitt]{kenton2023discovering}
Z.~Kenton, R.~Kumar, S.~Farquhar, J.~Richens, M.~MacDermott, and T.~Everitt.
\newblock Discovering agents.
\newblock \emph{Artificial Intelligence}, page 103963, 2023.

\bibitem[Klein(1967)]{klein1967glimpses}
O.~Klein.
\newblock Glimpses of {N}iels {B}ohr as scientist and thinker.
\newblock \emph{Niels Bohr. His life and work as seen by his friends and
  colleagues. London: Interscience Publishers}, pages 74--93, 1967.

\bibitem[Marr(2010)]{marr2010vision}
D.~Marr.
\newblock \emph{Vision: A computational investigation into the human
  representation and processing of visual information}.
\newblock MIT press, 2010.

\bibitem[Moreno(2018)]{moreno2018minimal}
A.~Moreno.
\newblock On minimal autonomous agency: natural and artificial.
\newblock \emph{Complex Systems}, 27\penalty0 (3), 2018.

\bibitem[Moreno and Etxeberria(2005)]{moreno2005agency}
A.~Moreno and A.~Etxeberria.
\newblock Agency in natural and artificial systems.
\newblock \emph{Artificial Life}, 11\penalty0 (1-2):\penalty0 161--175, 2005.

\bibitem[Ng and Russell(2000)]{ng2000algorithms}
A.~Y. Ng and S.~J. Russell.
\newblock Algorithms for inverse reinforcement learning.
\newblock In \emph{Proceedings of the International Conference on Machine
  Learning}, 2000.

\bibitem[Russell(1998)]{russell1998learning}
S.~Russell.
\newblock Learning agents for uncertain environments.
\newblock In \emph{Proceedings of the Conference on Computational Learning
  Theory}, 1998.

\bibitem[Tomasello(2022)]{tomasello2022evolution}
M.~Tomasello.
\newblock \emph{The evolution of agency: Behavioral organization from lizards
  to humans}.
\newblock MIT Press, 2022.

\bibitem[Zadeh(1963)]{zadeh1963definition}
L.~A. Zadeh.
\newblock On the definition of adaptivity.
\newblock \emph{Proceedings of the IEEE}, 51\penalty0 (3):\penalty0 469--470,
  1963.

\bibitem[Ziebart et~al.(2008)Ziebart, Maas, Bagnell, Dey,
  et~al.]{ziebart2008maximum}
B.~D. Ziebart, A.~L. Maas, J.~A. Bagnell, A.~K. Dey, et~al.
\newblock Maximum entropy inverse reinforcement learning.
\newblock In \emph{Proceedings of the AAAI Conference on Artificiall
  Intelligence}, 2008.

\end{thebibliography}

\end{document}